\documentclass[letterpaper, 10 pt, conference, twoside]{IEEEtran}
\usepackage{enumitem,kantlipsum}
\usepackage{subfig}
\usepackage{times}
\usepackage{epsfig}
\usepackage{graphicx}
\usepackage{amsmath}
\usepackage{amssymb}
\usepackage{times}
\usepackage{caption}
\usepackage{multirow}
\usepackage{xcolor}
\usepackage{mathtools}
\usepackage{physics}
\usepackage{stackengine}
\usepackage{enumitem}
\usepackage{lineno}
\usepackage{lipsum,etoolbox}
\usepackage{tabularx}
\usepackage{lineno}
 \usepackage{mathptmx}
 \usepackage[symbol]{footmisc}
 \usepackage{url}
 \usepackage{hyperref}
\usepackage{geometry}
\pdfoutput=1
\usepackage{geometry}
\ifCLASSINFOpdf
\else
\fi
\begin{document}
%
% paper title
% Titles are generally capitalized except for words such as a, an, and, as,
% at, but, by, for, in, nor, of, on, or, the, to and up, which are usually
% not capitalized unless they are the first or last word of the title.
% Linebreaks \\ can be used within to get better formatting as desired.
% Do not put math or special symbols in the title.
\title{Low to High Dimensional Modality Hallucination using Aggregated Fields of View}
%  \author{Kausic Gunasekar$^{1}$ , Qiang Qiu$^{2}$  and Yezhou Yang$^{1}$% <-this % stops a space

% 0\thanks{$^{1}$Arizona State University }
%         % University of Twente, 7500 AE Enschede, The Netherlands
%         % {\tt\small albert.author@papercept.net}}%
% \thanks{$^{2}$ Duke University}
% % Bernard D. Researcheris with the Department of Electrical Engineering, Wright State University,
% %         Dayton, OH 45435, USA
% %         {\tt\small b.d.researcher@ieee.org}}%
% }
% \author{Kausic Gunasekar$^{1}$ and Qiang Qiu$^{2}$ and Yezhou Yang$^{1}$% <-this % stops a space
% \thanks{$^{1}$Active Perception Group, CIDSE, Arizona State University }%
% \thanks{$^{2}$ECE, Duke University}%
% }

\author{\IEEEauthorblockN{Kausic Gunasekar\IEEEauthorrefmark{1},
Qiang Qiu\IEEEauthorrefmark{2}, and
Yezhou Yang\IEEEauthorrefmark{1}}
\IEEEauthorblockA{\IEEEauthorrefmark{1} Arizona State University, Arizona, USA,
Email : kgunase3@asu.edu, yz.yang@asu.edu}
\IEEEauthorblockA{\IEEEauthorrefmark{2}Duke University, North Carolina, USA,
Email: qiang.qiu@duke.edu}}%

\maketitle
\setlength{\belowcaptionskip}{-4pt}
\setlength{\abovedisplayskip}{-3pt}
\setlength{\belowdisplayskip}{-2pt}
% As a general rule, do not put math, special symbols or citations
% in the abstract or keywords.
\begin{abstract}
Real-world robotics systems deal with data from a multitude of modalities, especially for tasks such as navigation and recognition. The performance of those systems can drastically degrade when one or more modalities become inaccessible,  due to factors such as sensors' malfunctions or adverse environments. Here, we argue modality hallucination as one effective way to ensure consistent modality availability and thereby reduce unfavorable consequences. While hallucinating data from a modality with richer information, e.g., RGB to depth,  has been researched extensively, 
we investigate the more challenging low-to-high modality hallucination with interesting use cases in robotics and autonomous systems. We present a novel hallucination architecture that aggregates information from multiple fields of view of the local neighborhood to recover the lost information from the extant modality. The process is implemented by capturing a non-linear mapping between the data modalities and the learned mapping is used to aid the extant modality to mitigate the risk posed to the system in the adverse scenarios which involve modality loss. We also conduct extensive classification and segmentation experiments on UWRGBD and NYUD datasets and demonstrate that hallucination allays the negative effects of the modality loss. Implementation and models:
\url{https://github.com/kausic94/Hallucination}.
\end{abstract}

% Note that keywords are not normally used for peerreview papers.
% \begin{IEEEkeywords}
% IEEE, IEEEtran, journal, \LaTeX, paper, template.
% \end{IEEEkeywords}
% \begin{IEEEkeywords}
%  Deep Learning in Robotics and Automation; Robot Safety; Modality Hallucination; Computer Vision for Other Robotic Applications. 
% \end{IEEEkeywords}

% For peer review papers, you can put extra information on the cover
% page as needed:
% \ifCLASSOPTIONpeerreview
% \begin{center} \bfseries EDICS Category: 3-BBND \end{center}
% \fi
%
% For peerreview papers, this IEEEtran command inserts a page break and
% creates the second title. It will be ignored for other modes.
\IEEEpeerreviewmaketitle

\section{Introduction}
\begin{figure}[!t]
    \centering
    \includegraphics[width = \linewidth]{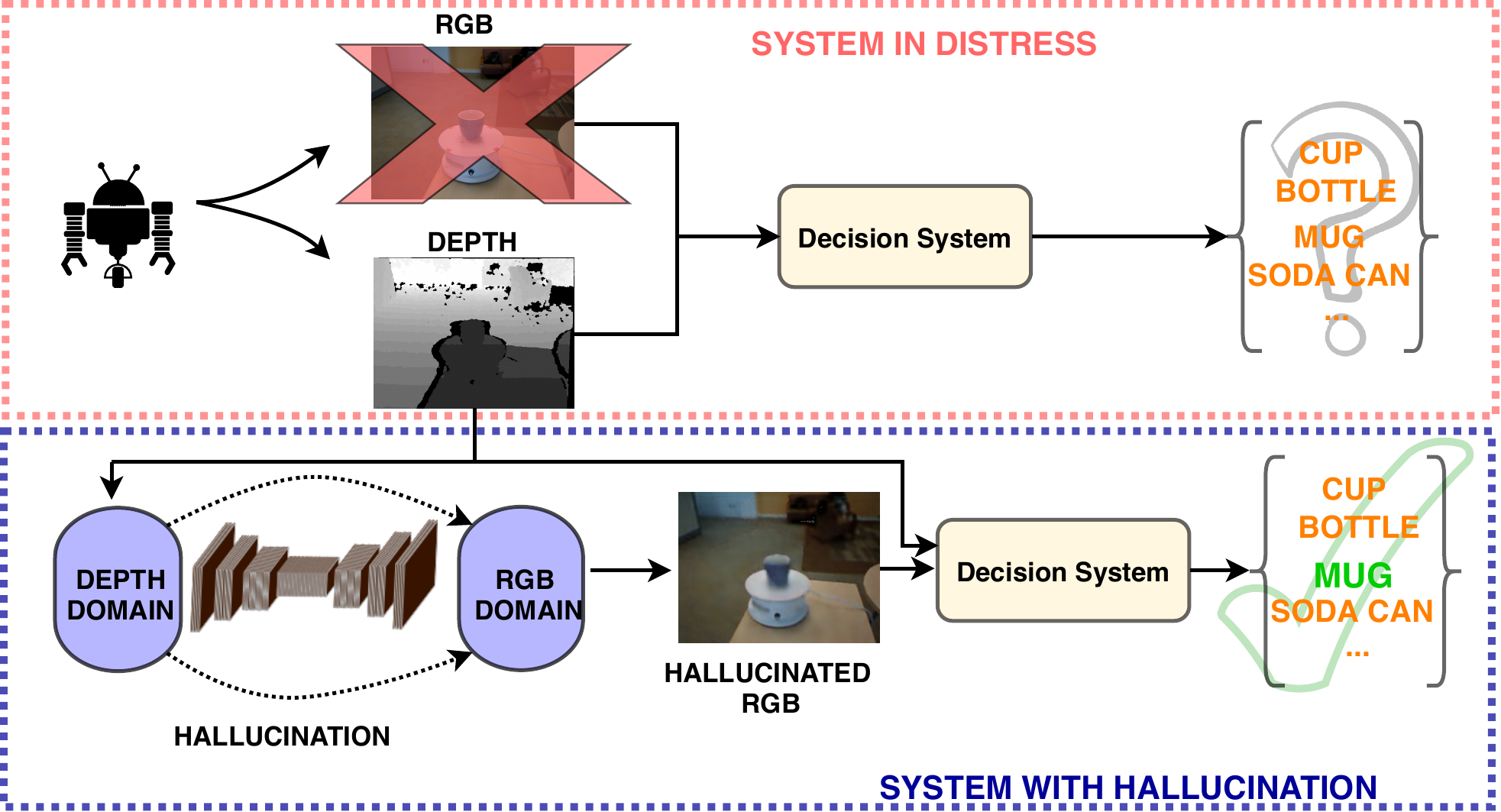}
    \caption{Illustration of a system that mitigates risk by adopting a hallucination during the adverse scenario of modality loss. Best viewed in color. }
    \label{illustration}
\end{figure}

% The very first letter is a 2 line initial drop letter followed
% by the rest of the first word in caps.
% 
% form to use if the first word consists of a single letter:
% \IEEEPARstart{A}{demo} file is ....
% 
% form to use if you need the single drop letter followed by
% normal text (unknown if ever used by the IEEE):
% \IEEEPARstart{A}{}demo file is ....
% 
% Some journals put the first two words in caps:
% \IEEEPARstart{T}{his demo} file is ....
% 
% Here we have the typical use of a "T" for an initial drop letter
% and "HIS" in caps to complete the first word.
% \IEEEPARstart{T}{his} demo file is intended to serve as a ``starter file''
% for IEEE journal papers produced under \LaTeX\ using
% IEEEtran.cls version 1.8b and later.
% You must have at least 2 lines in the paragraph with the drop letter
% (should never be an issue)
\IEEEPARstart{C}{ontemporary} robotic systems and intelligent agents such as autonomous ground or aerial vehicles, smartphones and nimble security systems heavily rely on processing information from multiple sensory data streams to yield accurate and reliable decision-making results. Usually, the systems are subjected to correlated data from numerous streams.  One best way to ensure the efficacy and efficiency of these systems in terms of performance is to add redundancy into the system by incorporating data from all possible streams. Recent advances in multimodal information fusion mechanisms  have made it possible to widely adopt these techniques and incorporate them within the systems to ensure the best performance. Thus, it would be ideal for a system to access as many modalities as possible. When the decision system makes use of all these different streams of data, it is a necessary precaution to have more than one sensor for each modality just in case one of them fails. In practice, however, various constraints like the system budget, physical form factor, power budget, etc make it problematic to integrate the required redundancy. One effective alternative that will help in beating the constraints and difficulties mentioned above is hallucinating the data of the desired modality from another modality. For example, predicting  the depth map of an image given its RGB image, with a trustworthy predicting method, we can replace the need of the redundant sensor with the predictor. An illustration of it is portrayed in Fig. \ref{illustration} 

%\setkeys{Gin}{draft=true}
For our study, we consider a particularly prevalent scenario in autonomous systems in which one or multiple sensors fail.  Although sensors can generally function reliably for a long period of time, the lingering risk still exists that certain channels of the sensor array may fail at a critical time. The notorious case, where an autonomous car hit a pedestrian recently happened in Arizona US \cite{news}, there is  speculation that the LIDAR sensor on the vehicle failed to function before the tragedy actually happened, and it is believed to be one of the crucial factors that caused the accident. So, as the hardware-level sensor malfunctioning is inevitable, are there any backup approaches an intelligent system can take to mitigate the risks or lower the likelihood of failure? More importantly, how to utilize the sensor channel with less information to hallucinate the information-rich sensor channel? Here, we put forward the first approach, to the best of our knowledge, that increases the reliability of a decision system involving multi-modal data. Especially, we consider a system that takes in two channels of sensory inputs: an RGB image channel and a depth channel. In our scenario, both channels are generating sensory data to the intelligent system while they are both functioning normally. We consider this data as the training data. In an adverse scenario wherein either of the channels fails to function, we want our system to be able to still function with a hallucinated backup channel. \newgeometry{top=54pt,left=54pt,right =54pt,bottom = 54pt} While  RGB to Depth has been widely studied under the topic of  ``single camera/monocular depth estimation''  \cite{depthprediction,depthprediction2}, we \noindent put our main focus of this paper onto the less explored scenario that the information-rich RGB  channel stops working, and needs to be recovered.  

Hallucinating from a lower information space to a higher information space has many challenges associated with capturing the right information. The capturing of this information should enhance the performance of the system which takes in this data. This is further constrained by the fact that we are aiming at a task-agnostic approach as well as an online trainable method, making the problem harder.

% Here put our take.

We summarize our contributions as follows:
\begin{enumerate}[noitemsep,nolistsep,leftmargin=*]

%\item We investigate state of the art discriminative and generative neural architectures for the purpose of hallucination and we treat them as our baseline.

\item We introduce a novel AggConv and AggTrConv block and designed a  encoder-decoder based neural network architecture  that incorporates multiple fields of view into its base encoder and decoder blocks to learn the low-to-high dimensional modality mapping. We compare the results of our architecture with that of the baseline results.

\item We design and conduct experiments on the well-known UWRGBD and the NYUD datasets, and empirically show the advantage in hallucinating with our architecture over the baseline architectures. We further validate the usefulness of the hallucinated data by subjecting the hallucinated data to two fundamental vision task, namely, image classification and semantic segmentation.

\item We also experimentally show an added advantage that we observed with the hallucinated data. By incorporating the hallucinated data into the original system it can further improve the performance of the original system.
\end{enumerate}

\

\section{RELATED WORK}

\noindent\textbf{Modality hallucination related:}
    Learning combined space representations and hallucinating data from different modalities is an active field of research \cite{qiupaper,learningside,unseen,dbm} as it provides many advantages to the system that incorporates it. The work done in \cite{qiupaper} is a very good example where the authors hallucinate RGB versions of the image from infra-red images and that is used in the face verification task. This, however, is targeted at a domain adaptation setting where the face verification model trained for RGB images is adapted to near infra-red images. The work by Hoffman et al. in \cite{learningside} also deals with modality hallucination, although their method is used to learn mid-level abstractions and that is further used to enhance the performance of the detection network. The learning of their hallucination network is embedded as part of their object detection module and it learns by loss function paired with the depth stream. They do not produce hallucinated data and are also restricted to a specific task. Other works have used a mapping between modalities to help better the performance of a system \cite{unseen} or use a generative model to learn the distribution of modalities and sample from them when needed \cite{dbm}. In \cite{unseen}, the authors learned a mapping using the unlabeled data with the help of Gaussian processes and that is leveraged for the object recognition task of objects previously not seen in the training data.  Their scenario is quite different as they propose to tackle missing data instances for their task while our work is focused on tackling missing modality using data hallucinated with the help of CNNs. The work done in \cite{dbm} uses generative models, specifically deep Boltzmann machines to help with the missing data. Generative models come with their disadvantages and an important one is that they do not produce a one-to-one mapping as required in these tasks, instead, it learns the distribution which may not well describe our missing data modality. Our work is substantially different as ours is not a generative model and we tackle the case of an entire modality missing and not missing instances of data. Other generative networks have also been used for similar tasks like in \cite{depthvae} and \cite{pixel} but differ in many ways from our work. They are not concerned with risk mitigation in system with a lost modality. \cite{depthvae} deals in producing a semantic map from a depth input image using variational autoencoders. RGB images (our case) belong to a higher dimensional modality and is much more information rich compared to semantic maps. \cite{pixel} targets at data augmentation for imbalanced segmentation data. Moreover, they use a 2-stage pipeline which means it cannot be adapted as online trainable model unlike ours. They use pix2pix \cite{pix2pix} as well for generating RGB from semantic labels and we empirically show that it doesn't work well in our case. {Work done in \cite{inpaint} and \cite{pixelaware} aims to solve different problem statements but is related to our work with regards to methodology and motivation. %Although the architectures bears certain similarities to ours 
    They are functionally different from ours as the features obtained from different kernels are fused only at the end. Our architecture enables complex intermixing of features from different kernels including the low level features thus utilizing contextual information at every level. }%Unlike, the work mentioned above our methodology is motivated by utilizing the correlation a feature has with it's neighborhood.}
    
\noindent\textbf{Multi-modal information processing:}
Multi-modal systems are becoming more common recently and consequently, there has been increased interest in this field of research \cite {receipe,multimedia,zeroshot,dist}.
%embeddings,crossmodalrep}
Work such as \cite{zeroshot,dist} deal with the learning of cross-modal data but differ in their learning process in the sense that they do not hallucinate the data in any manner similar to ours. While the former transfers images to the semantic text space and uses it to help in classification the latter uses a learned model to transfer its learning for the same task on the other modality. \cite{pedestrian}  deals with RGB and thermal data modalities for the case of pedestrian detection. But in their work, they are using RGB images to reconstruct thermal images (high to low dimensional mapping) and using them on their detection network. Unlike \cite{pedestrian} our hallucination scheme is generic and not task-specific. A lot of work has been done that deals with multi-modal information processing. Work done in \cite{receipe,multimedia} involves learning cross-modal representations and associating the modality embeddings to learn the relationship between the modalities. This learned information is used to perform a specific task on a given data modality as in \cite{receipe}.

\section{ Network Architecture: }
We propose a network architecture that is specifically designed to utilize more neighborhood information of the current feature in building the next layer of feature maps. Specifically, we introduce the AggConv and AggTrConv blocks as the base building blocks and we construct our architecture using those blocks. Traditionally convolutional neural networks work on a single sized kernel which restricts the field of view, to that kernel.As prediction from a lower to higher dimensional modality is fundamentally an under-constrained problem, it becomes increasingly hard for neural networks to find the right solution with the conventional small field of view from a single kernel. To tackle this, additional information from the local neighborhood of the pixel can be utilized to help the network in hallucinating better. Thus, we believe, a  hallucination network that uses information obtained from different fields of view will best suit this ill-posed problem. Different fields of view encapsulate various degrees of information to pass on to the next set of feature maps. In the case of just a single field of view, in the lower dimensional modality, the kernel cannot capture information that is sufficient for reconstruction in the higher dimension. But by aggregating the different fields of view and thus expanding the same, the network can leverage the relationship that exists between the pixels and their neighbors to predict correctly in the higher dimensional space.

\begin{table}[!ht]
    \centering
    \resizebox{0.9\columnwidth}{!}{
    \begin{tabular}{|c|c|c|c|c|}
    \hline
    \multicolumn{5}{|c|}{\textbf{Network Architecture}} \\
    \hline
    \hline
         Name & Layer & Filters & Skip & Kernels    \\
    \hline 
    Enc\_1 & Encoder & 48 & - & - \\
    Enc\_2 & Encoder & 60 & - & - \\
    Enc\_3 & Encoder & 192 & - & - \\
    Enc\_4 & Encoder & 288 & - & - \\
    Dec\_4 & Decoder & 96 & - & - \\
    Dec\_3 & Decoder & 30 & Enc\_3 & - \\
    Dec\_2 & Decoder & 24 & Enc\_2 & - \\ 
    Dec\_1 & Decoder & 3 & Enc\_1 & - \\
    logits & Convolution & 3 & - & 5x5 \\
    Hallucinated & Convolution & 3 & - & 3x3 \\
    \hline
    \end{tabular}}
    \caption{This table describes the complete  architecture used in our experiments. }
    \label{myarch2}
    
\end{table}

{ The AggConv and AggTrConv blocks serve in expanding the field of view by concatenating features from different fields of view. The AggConv and AggTrConv blocks have several advantages which make them ideal for this purpose. The AggConv blocks utilize multiple kernels of different sizes and accumulate the feature maps produced by each kernel. Kernels with a bigger field of view are used with the help of dilated convolutions. This ensures that AggConv blocks are not memory intensive which would be the case if bigger un-dilated kernels were used. For instance, using the 11x11 kernel with a dilation size of 3 effectively covers a neighborhood of 31x31. By using this instead of 31x31 kernel saves about 87\% of parameters in this case. The features used are sparser compared to the un-dilated kernel but by using a combination of such kernels the network obtains the information it needs. The AggConv block has a convolution operation with just a 3x3 kernel to sub-sample the feature maps when it is needed. While the AggConv blocks are used for downsampling the feature maps into a feature-rich latent space a similar AggTrConv block is used to up-sample and eventually reconstruct the lost modality. The design of our AggConv and AggTrConv blocks are further explained in table \ref{myarch}. Another advantage of these blocks that we introduced is that, they can be easily parallelized as it involves multiple operations on a single input which paves way for concurrency, thus ensuring faster compute than what would be expected in an architecture that is operation heavy.  The architecture is built in an encoder-decoder fashion as shown in table \ref{myarch2}. The encoder and decoder blocks are built with AggConv and AggTrConv blocks as shown in Table \ref{myarch}. The architecture is defined with 4 encoder and 4 decoder blocks. Skip connections are used to add the activation output from each encoder layer to the corresponding decoder layer. The network takes in depth images as input and RGB images as the ground truth data. }

    \begin{table}[!ht]
        \centering
        % \subfloat{
        \vspace{9pt}
        \resizebox{0.8\columnwidth}{!}{
        \begin{tabular}{|c|c|c|c|}
        \hline
        \multicolumn{4}{|c|}{\textbf{Encoder - Decoder Blocks}} \\
        \hline
        \hline
         Layer & Filters & Stride & Skip Connection \\
        \hline
        \multicolumn{4}{|c|}{\textbf{Encoder block} :  Filters : d}\\
        \hline
        AggConv & d & 1 & NO  \\
        \hline
        ReLU & - & -  & -  \\
        \hline
        AggConv & d & 1 & YES \\
        \hline
        ReLU & - & - & - \\
        \hline
        Convolution & \multirow{3}{*}{d} & \multirow{3}{*}{2} & \multirow{3}{*}{NO} \\
        \textit{kernel : 3x3} & & & \\
        \textit{dilation rate : 1} & & & \\
        \hline
        \multicolumn{4}{|c|}{\textbf{Decoder block} :  Filters : d}\\
        \hline
        AggTrConv & d & 0.5 & NO \\
        \hline
        ReLU & - & - & - \\
        \hline
        AggTrConv & d & 1 & YES \\
        \hline
        ReLU & - & - & - \\
        \hline
        \end{tabular}
        }
        % }}
        % \subfloat{
        \bigskip
        
        \resizebox{\columnwidth}{!}{
        \begin{tabular}{|c|c|c|c|c|}
        \hline
         \multicolumn{5}{|c|}{\textbf{Architecture Building Blocks}} \\
         \hline
         \hline
         Layer & Kernels & Filters & dilation rate & stride \\
         \hline
         \multicolumn{5}{|c|}{\textbf{AggConv Block} :  Filters : d , Stride : s}\\
         \hline
         convolution & 3 x 3 & d/6 & 1 & s \\
                    & 11 x 11 & d/6 & 1 & s \\
                    & 5 x 5 & d/6 & 2 & s  \\
                    & 7 x 7 & d/6 & 2 & s \\
                    & 9 x 9 & d/6 & 3 & s \\
                    & 11 x 11 & d/6 & 3 & s \\
        \hline
        concatenation & - & - & - & - \\
        \hline
        batch Normalization & - &  - & - & - \\
        \hline 
        \multicolumn{5}{|c|}{\textbf{AggTrConv Block} :  Filters : d , Stride : s}\\
        \hline
        convolution & 3 x 3 & d/3 & 1 & s \\
                    & 7 x 7 & d/3 & 1 & s \\
                    & 11 x 11 & d/3 & 1 & s \\
        \hline
        concatenation & - & - & - & - \\
        \hline 
        batch Normalization  & - & - & - & - \\
        \hline
        \end{tabular}}
        % }}
        \caption{Encoder - Decoder blocks constructed using the AggConv and AggTrConv blocks. This table describes the basic building blocks that is used in the architecture in table}
        
        \label{myarch}
    \end{table}

\section {Baseline and Validation Networks}
To best understand the superior ability of our network in hallucination we compare it to baseline models obtained by subjecting the hallucination procedure with networks that have proven well in image to image prediction tasks. The process of hallucination is formulated as an image translation problem from the depth to the RGB domain and we use a conditional GAN \cite{pix2pix}, popularly known as pix2pix which has been shown to do well in the task of image translation. We also use an off-the-shelf semantic segmentation network \cite{linknet} called linkNet and re-purpose it for the task of hallucination. Linknet is also an encoder-decoder architecture which has been shown to outperform networks like deeplab\cite{deeplab} in a parameter efficient way. 

 The effectiveness of the reconstructed modality cannot be judged visually. We judge it by subjecting the reconstructed modality in central vision tasks of object classification and semantic segmentation. The hallucinated images produced by our network and the baseline networks are compared in the above tasks and we report the numbers for the same. We use AlexNet \cite{alexnetpaper} for the classification task and \cite{seg} for the semantic segmentation task. 
\section{Loss Function}

The hallucinator loss $\mathcal{L}_{hal}$ is formulated with two components. They are the root mean square loss term and the smoothness constraint. Here, $\lambda$ is used to control the relative importance of the different loss functions. Shown in Eq.~\ref{final} is a pixel-wise loss formulation:

\begin{equation}
\mathcal{L}_{hal}  = \mathcal{L}_{rmse} + \lambda \mathcal{L}_{smooth}.
\label{final}
\end{equation}
A root mean squared error between the hallucinated images produced by the hallucination model and the ground truth images obtained from RGB cameras helps to capture the important abstraction between the two spaces. The main goal of this hallucination network is to capture the non-linear relationship between the spaces.
The Eq.~\ref{rmse} works well to capture the said abstraction.

\begin{align}
  \mathcal{L}_{rmse}  &= \sqrt{\frac{\sum_{i=1}^{N}(p_{i} - \bar{p}_i)^2}{N}}, 
\label{rmse}
\end{align}
where $N$ represents the number of pixels in target image $I$, and $p_{i}$, $\bar{p}_{i}$ the ground truth and reconstructed pixel respectively.
To obtain a consistent mapping and to lessen haphazard or chaotic results we introduce an edge aware smoothness constraint. Smoothness constraints are commonly used in depth prediction like the work in  \cite{dense, monocular}. The smoothness constraint should enforce local smoothness and at the same time should preserve the edges, formulated in the Eq.~\ref{smooth}:
\vspace{-5pt}
\begin{equation} 
 \mathcal{L}_{smooth} =\frac {1}{N}\sum_{N}{\textit{H}}(\triangledown I_{hal}) e^{-{\textit{H}}(   \triangledown I) }, 
\label{smooth}
\end{equation}
where $I_{hal}$ represents the hallucinated tensor, ${I}$ the ground truth, and {\textit{H}} is the \textit{Huber} function~\cite{huber1964robust}:

 \begin{equation}
     \textit{H}(x) =
    \begin{cases}
      \frac{1}{2}x^{2} & \text{ if $\lvert x \rvert \leq \delta $ }\\
      \delta(\lvert x \rvert - \tfrac{1}{2}\delta) & \text{otherwise}
    \end{cases},
 \end{equation}
where $\delta$ = $\abs{x}$, and $N$ the total number of pixels within one training batch.
 
%  All the hallucination models other than the GANs are trained with the formulation shown in eqn. \ref{final} The GANs are trained with the loss function mentioned in \cite{pix2pix}. The classification and segmentation networks are subjected to the standard softmax cross entropy loss as shown in Eq.~\ref{ce}. 
%  \begin{equation}
% \centering
% \begin{aligned}
%  \mathcal{L}_{ce} = -\sum_{i=1}^{C} \bar{y}_{i} \log{y_i},
% \end{aligned}
% \label{ce}
% \end{equation}
% where $\mathcal{L}_{ce}$ is the  cross entropy loss for the task and $C$ is the number of classes $\bar{y}_i$ represents the ground truth annotation while ${y}_i$ is the predicted output. For classification the loss computed is between the prediction distribution and the class labels. For segmentation it is a pixel-wise classification between the annotated map and the predicted one. 

\section{Experiments}
%The dataset for the baseline models as well as our proposed architecture is maintained the same. 
% {\color{red} We demonstrate our method on Depth and RGB domain pairs as they are cleanly labeled for particular tasks (classification and segmentation) and in bulk numbers.} %YZ: no need in my view. 
The same set of input depth images are used in training of all the models and they are all tested on the same test set which is seperated  from the training set to ensure fairness. A similar procedure is followed for classification and segmentation experiments.

\subsection{Dataset}
 
\textbf{Hallucination:} We design our experiments on the following two datasets:  NYUD dataset \cite{nyud1} and the University of Washington's RGBD dataset \cite{uwrgbd}. The datasets mentioned above have RGB images and their corresponding depth images. The UWRGBD dataset has over 200,000 images belonging to 51 classes. Although the UWRGBD dataset has over 200K images the dataset is heavily skewed. For instance, some classes have less than 2000 images while others have over 10,000 images. To make sure there is no untoward bias, the dataset is split into 875 images per class for training and 100 images per class for testing. Hence, in total 44,625 training images and 5100 testing images are obtained for the hallucination experiment.  The NYUD -v1 dataset, on the other hand, has only 2284 labeled images. So we used the raw dataset available in the NYUD-V1 \cite{nyud1}. The raw dataset, unlike the labeled dataset, contains depth images that are not in-painted along with their corresponding RGB images and there are over 100000 such pairs.  The NYUD V1 dataset has in total of 135,314 RGB - depth image pairs. The raw depth images are in-painted to remove artifacts using a cross bilateral filter\cite{cbf} and then projected onto the RGB plane and linearly scaled to get the depth image representation. The images were split into train and test set with an 80:20 ratio. 

\textbf{Classification: } For classification, we used 500 images for training and 175 images for testing per class from the UWRGBD dataset. Thus in total 25,500 training images and 8925 testing images are used for classification. None of these training and testing images overlap with the hallucination dataset.  The dataset was subjected to a 51-way classification task. The images are cropped to only include the object to ensure there is no untoward data leakage.

\textbf{Segmentation: } The segmentation experiment was carried out with NYUD-v1 dataset that has 2284 labels from 64 different indoor scenes. The depth images are in-painted to fill holes just like it was done for the Hallucination dataset. The dataset was split into 70:30  training-testing split. The segmentation task was 40 class segmentation procedure. The hallucinated network trained on NYUD-v1 raw images was used to obtain the hallucinated images here.

\subsection{Implementation Details}
The hallucination experiments are done with the images maintained in their standard size of 640x480. The hallucination experiments were carried out in the YUV colorspace. The hallucinations using Aggregated convolution block architecture is implemented as a multi-GPU training pipeline. For both the NYUD dataset hallucination as well UWRGBD hallucination the experiments were carried out with ADAM optimizer with learning rate = 0.0005 , $\beta_1$ = 0.9 , $\beta_2$ = 0.999, $\epsilon$=1e-08. The Huber delta in the loss function \ref{smooth} ($\delta$) was set to 0.001 and the smoothing weight $\lambda$ is set to 50 The architecture was implemented using data parallelism on 3 GPUs. A total batch size of 21 is used in training with each GPU taking in 7 images per batch. The training for classification was carried out with a batch size of 200 and a learning rate of 1e-05 for 5 epochs. The semantic segmentation was completed with a learning rate of 1e-05 and a batch size of 25 for 100,000 steps. Both classification and segmentation are subjected to cross-entropy loss.

\subsection{Results}
\setkeys{Gin}{draft=false}
\subsubsection{GAN results}
\begin{figure}[!ht]
\centering
\includegraphics[width=.32\linewidth]{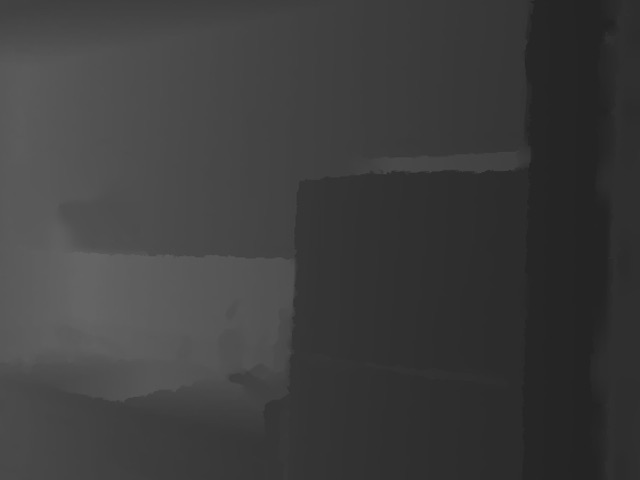}
\includegraphics[width=.32\linewidth]{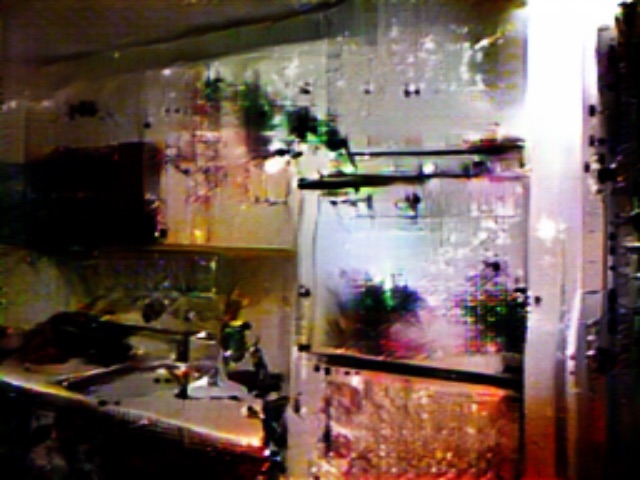}
\includegraphics[width=.32\linewidth]{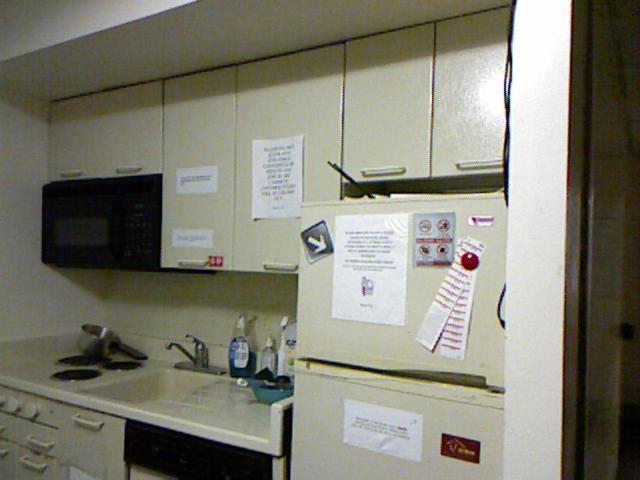}
\smallskip

% \includegraphics[width=.32\linewidth]{GANHalResults/resized_depth_0105.jpg}
% \includegraphics[width=.32\linewidth]{GANHalResults/resized_hallucinatedGAN_depth_0105.jpg}
% \includegraphics[width=.32\linewidth]{GANHalResults/resized_rgb_0105.jpg}
% \smallskip

\stackunder[3pt]{\includegraphics[width=0.32\linewidth]{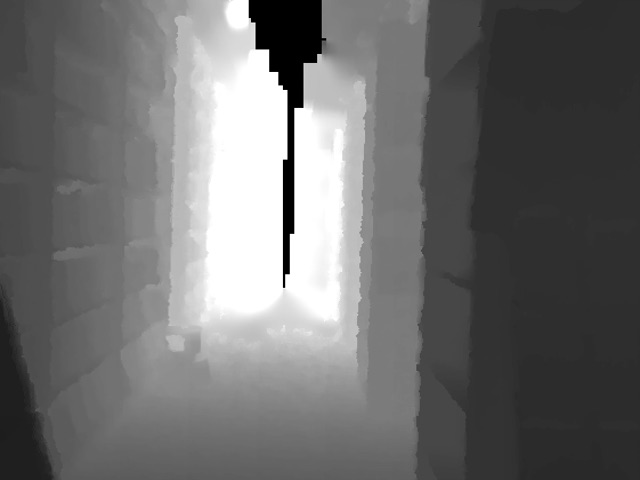}}
{(a) Depth}
\stackunder[3pt]{\includegraphics[width=.32\linewidth]{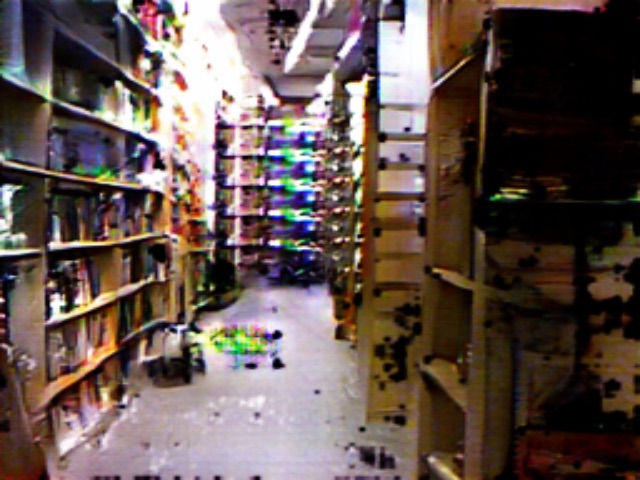}}{(b) Hallucinated}
\stackunder[3pt]{\includegraphics[width=.32\linewidth]{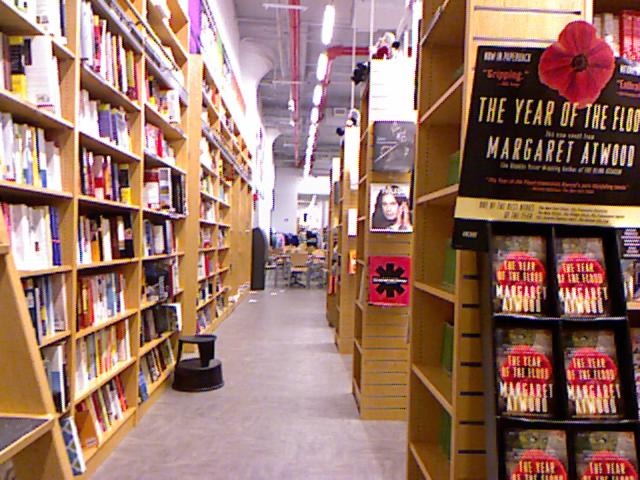}}{(c) RGB}
\caption{NYUD dataset hallucination results with GANs. (a) depth input image, (b) hallucination results,(c) groundtruth. }
\label{ganhalnyud}
\end{figure}

The generative model based hallucination procedure explained in the above sections with the pix2pix architecture does a relatively average job. Although it produces a few prominent structures with very little variance in the dataset like the turn-table in the UWRGBD dataset. The same pattern can be observed in the NYUD dataset as well. Moreover, the texture of the image compared to other those hallucinated by the Aggregated and linkNet architectures is less accurate. Some examples from the test set of the GAN based hallucination are shown in Fig. \ref{ganhalnyud}.
\begin{figure}[!ht]
\centering
\includegraphics[width=.32\linewidth]{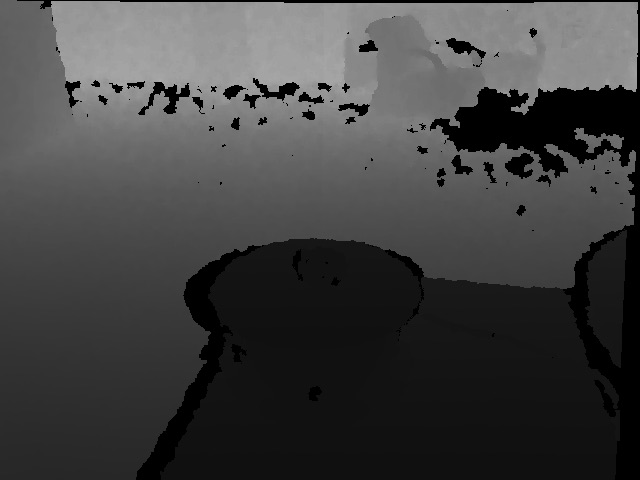}
\includegraphics[width=.32\linewidth]{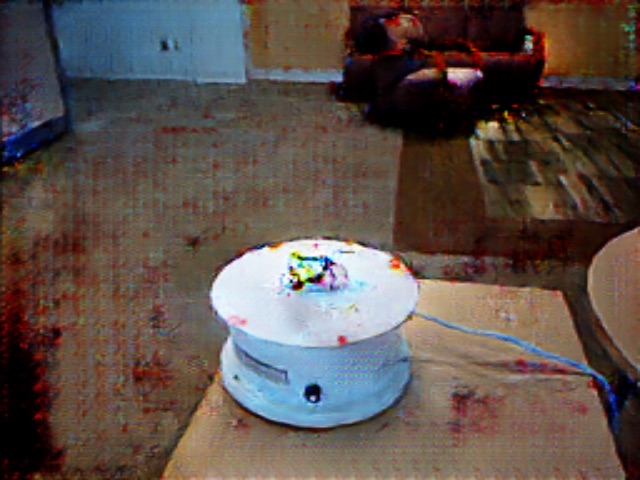}
\includegraphics[width=.32\linewidth]{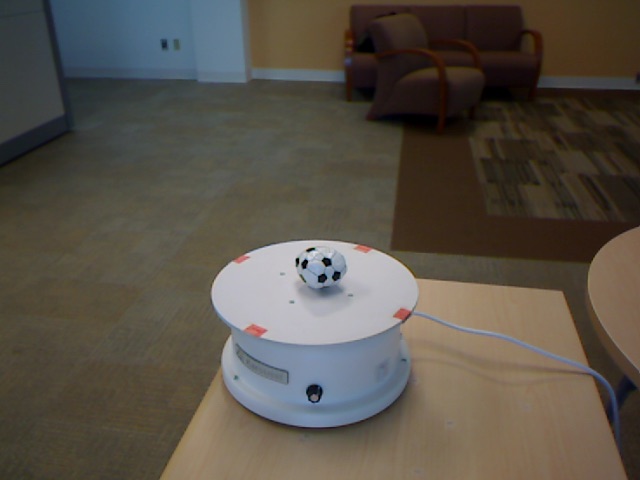}
\smallskip

\stackunder[3pt]{\includegraphics[width=0.32\linewidth]{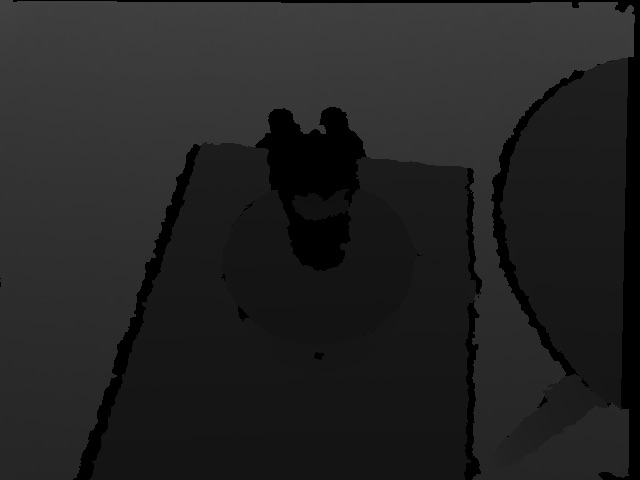}}
{(a) Depth}
\stackunder[3pt]{\includegraphics[width=.32\linewidth]{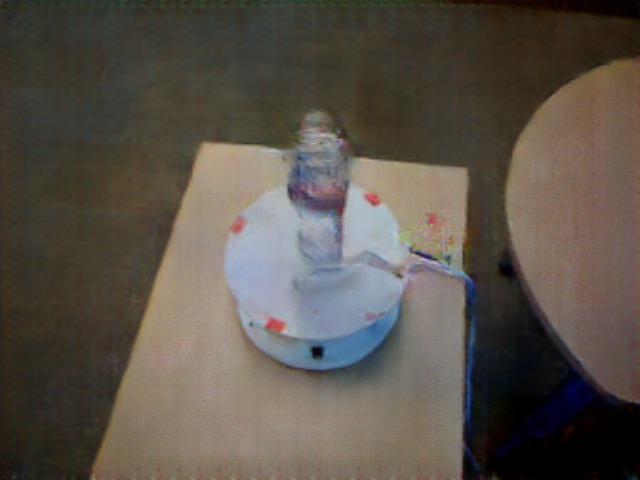}}{(b) Hallucinated}
\stackunder[3pt]{\includegraphics[width=.32\linewidth]{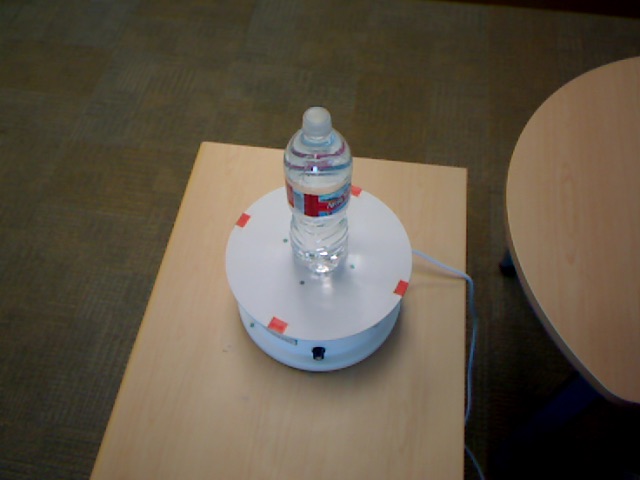}}{(c) RGB}
\caption{UWRGBD dataset hallucination results using the GANs (pix2pix architecture). (a)  depth input image, (b) hallucination results ,(c) is groundtruth. }
\label{ganhaluw}
\end{figure}
However, when it comes to reconstructing the pixels of the object of interest, it doesn't work as well. The pix2pix network retains a little bit of the structure of the object of interest but it doesn't produce a qualitatively well-defined structure. In many cases, it completely misses the color and structure of the object even in a relatively easy dataset like the UWRGBD dataset. Some results from the UWRGBD dataset hallucinated using GANs can be seen in \ref{ganhaluw}.

\subsubsection{LinkNet Results}
The linkNet architecture has an encoder-decoder architecture that compresses the image to a smaller dimension representation and reconstructs from that smaller dimension space. The linkNet architecture does a much better job of learning the correct family of functions for the mapping to take place and does so in a parameter efficient manner.  Fig.\ref{linknetHalUW} depicts results on the UWRGBD dataset.
\newcommand{\figscale}{.24\linewidth}
\begin{figure}[!h]
\centering
\includegraphics[width=\figscale]{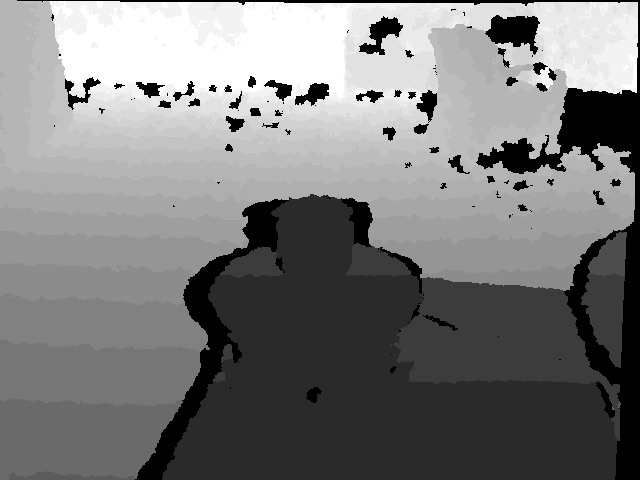}
\includegraphics[width=\figscale]{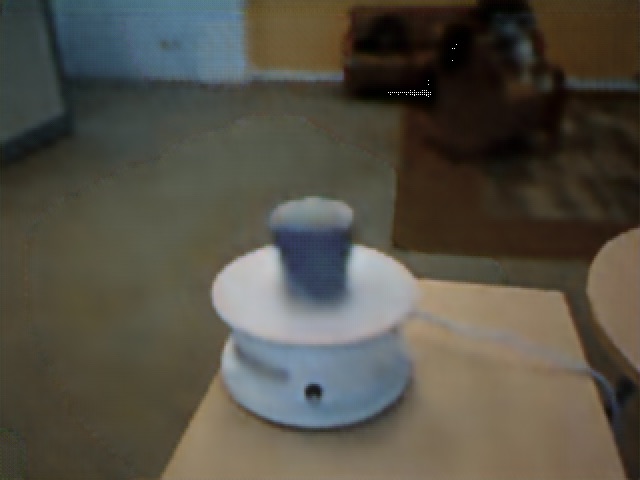}
\includegraphics[width=\figscale]{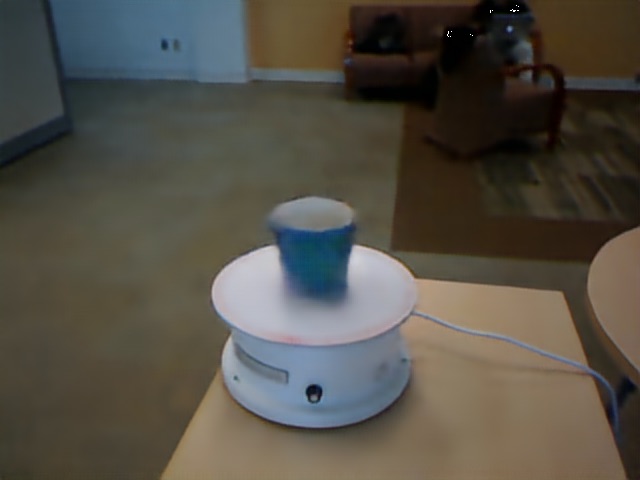}
\includegraphics[width=\figscale]{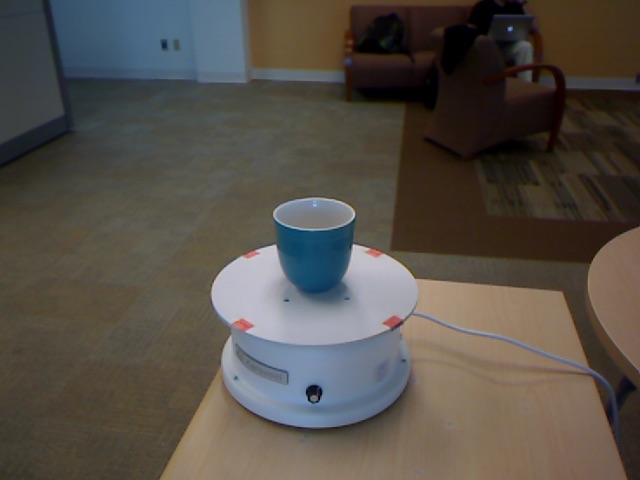}\smallskip

\stackunder[3pt]{\includegraphics[width=\figscale]{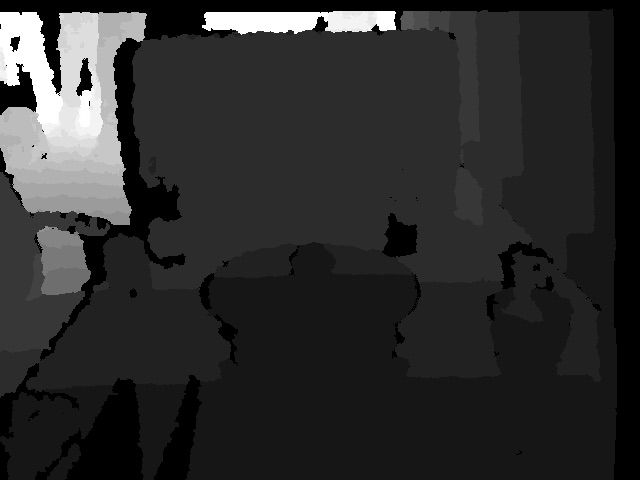}}
{(a) Depth}
\stackunder[3pt]{\includegraphics[width=\figscale]{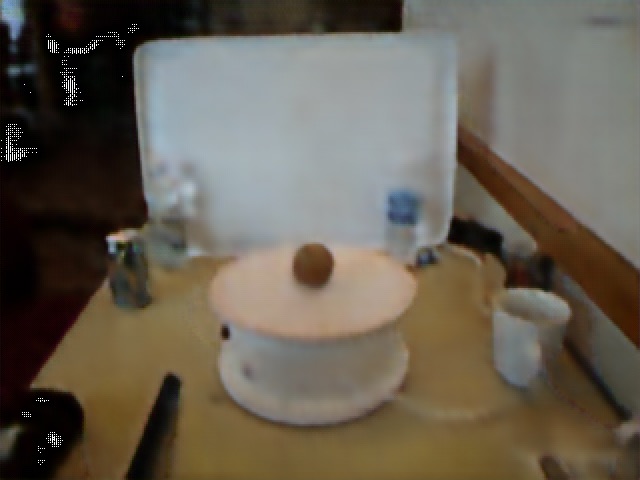}}{(b) Hal}
\stackunder[3pt]{\includegraphics[width=\figscale]{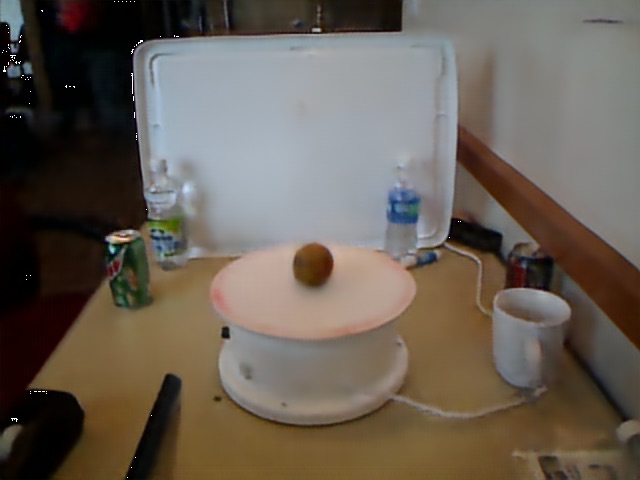}}{(c) Hal*}
\stackunder[3pt]{\includegraphics[width=\figscale]{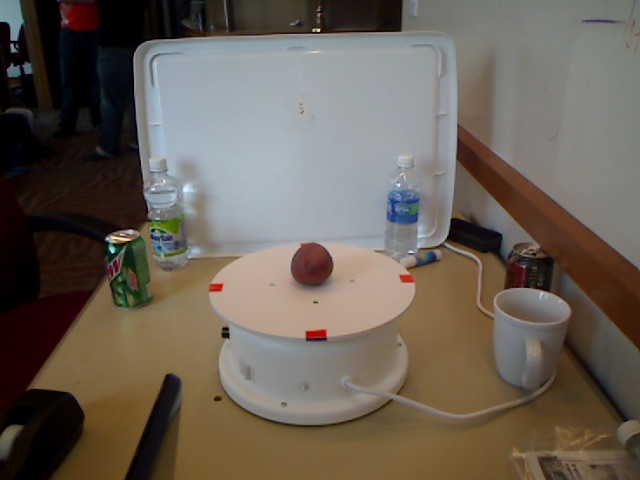}}{(d) RGB}
\caption{UWRGBD dataset hallucination results using the linkNet architecture. (a) depth input image, (b) hallucination result after first stage, (c) hallucination result after regularizing stage, (d) is groundtruth. }
\label{linknetHalUW}
\end{figure}

\begin{figure}[!h]
\centering
\includegraphics[width=\figscale]{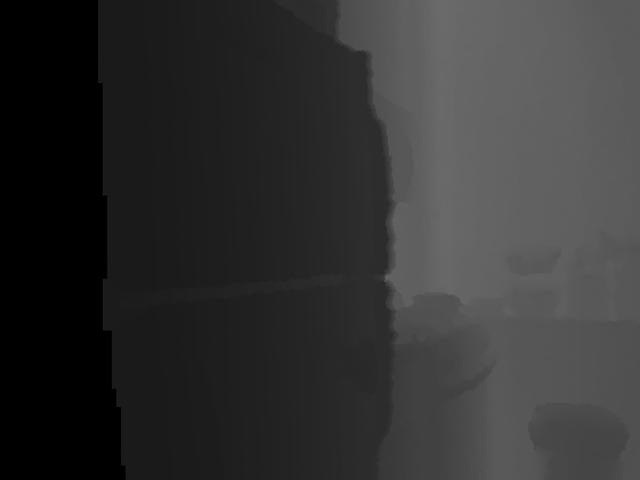}
\includegraphics[width=\figscale]{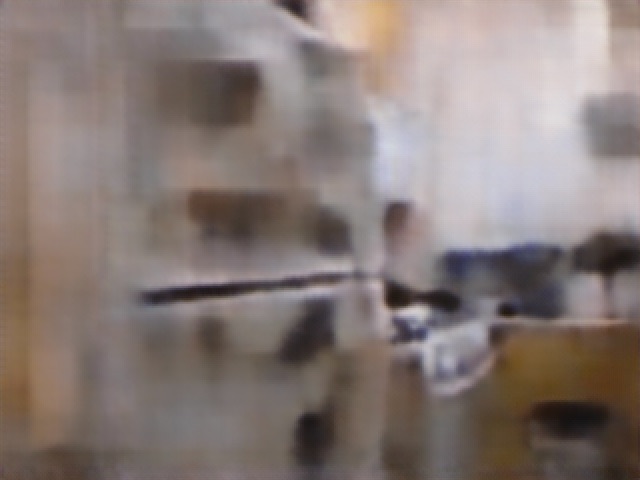}
\includegraphics[width=\figscale]{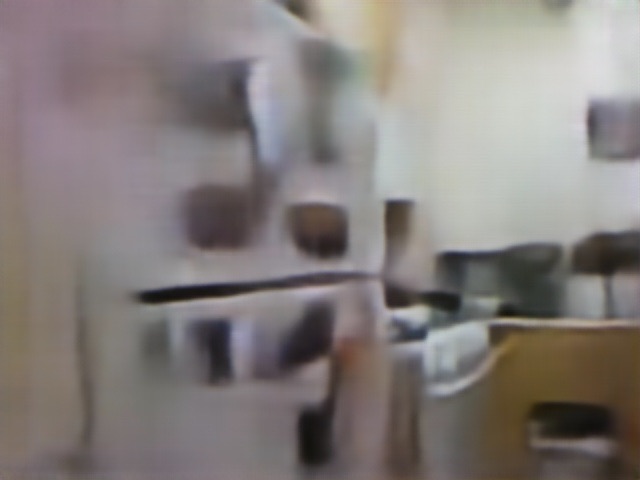}
\includegraphics[width=\figscale]{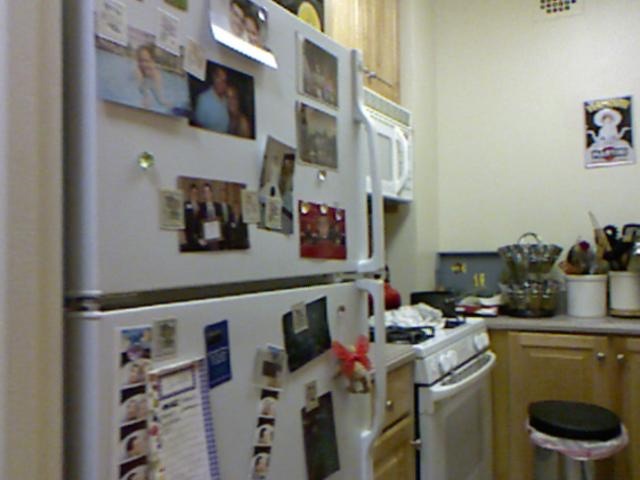}\smallskip

\stackunder[3pt]{\includegraphics[width=\figscale]{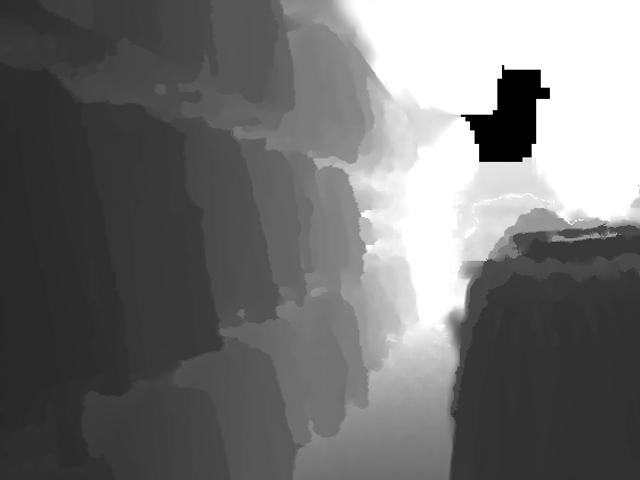}}
{(a) Depth}
\stackunder[3pt]{\includegraphics[width=\figscale]{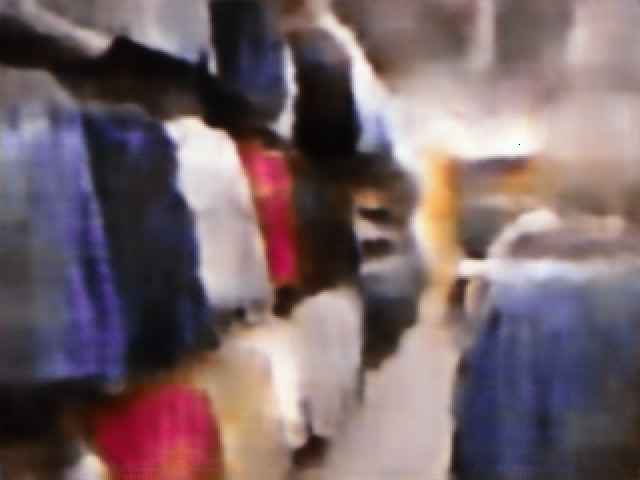}}{(b) Hal}
\stackunder[3pt]{\includegraphics[width=\figscale]{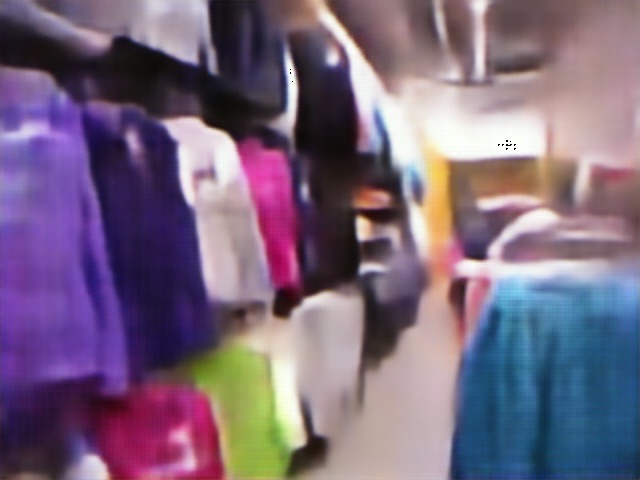}}{(c) Hal*}
\stackunder[3pt]{\includegraphics[width=\figscale]{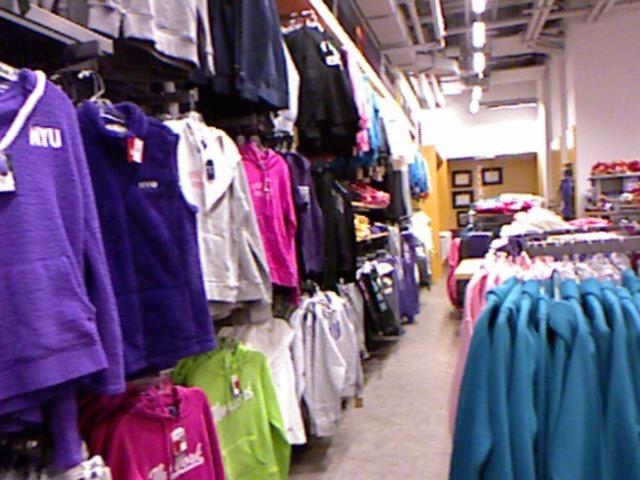}}{(d) RGB}
\caption{NYUD hallucination results using the linkNet architecture. (a) depth input image, (b) hallucination result  after the first stage, (c) hallucination result after regularizing stage, (d) is groundtruth. }
\label{linknetHalNYUD}
\end{figure}

As the UWRGBD is an easy dataset the hallucination does a pretty good job in the first stage itself but we still can find minor improvements after the regularization stage.
The NYUD dataset, on the other hand, provides a lot of evidence for the effectiveness of the regularization stage. The results from the NYUD dataset can be seen in Fig.\ref{linknetHalNYUD}.

\begin{figure}[!htb]

\minipage{0.33\linewidth}
  \includegraphics[width=\linewidth]{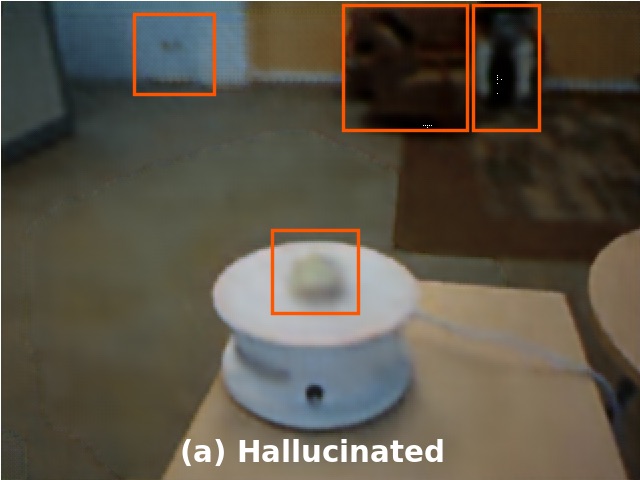}\smallskip
\endminipage\hfill
\minipage{0.33\linewidth}
  \includegraphics[width=\linewidth]{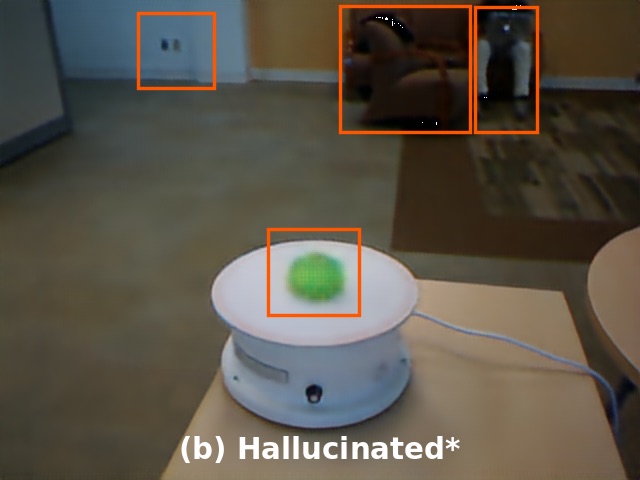}\smallskip
\endminipage\hfill
\minipage{0.33\linewidth}%
  \includegraphics[width=\linewidth]{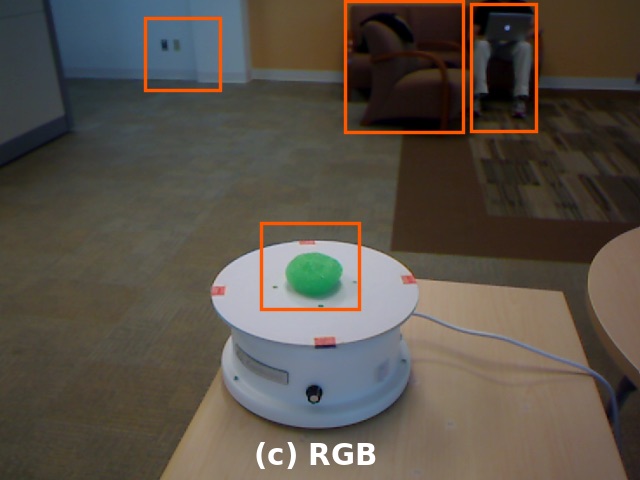}\smallskip
\endminipage

\minipage{0.245\linewidth}

  \includegraphics[width=\linewidth]{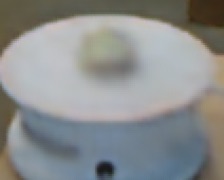}\smallskip
\endminipage\hfill
\minipage{0.245\linewidth}
  \includegraphics[width=\linewidth]{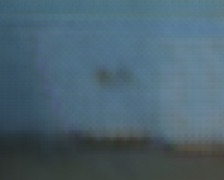}\smallskip
\endminipage\hfill
\minipage{0.245\linewidth}
  \includegraphics[width=\linewidth]{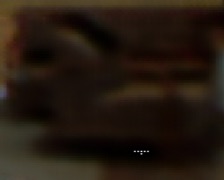}\smallskip
\endminipage\hfill
\minipage{0.245\linewidth}
  \includegraphics[width=\linewidth]{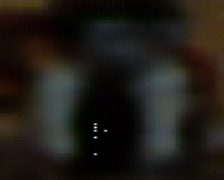}\smallskip
\endminipage\hfill

\minipage{0.245\linewidth}

  \includegraphics[width=\linewidth]{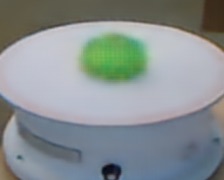}\smallskip
\endminipage\hfill
\minipage{0.245\linewidth}
  \includegraphics[width=\linewidth]{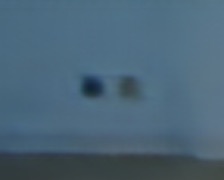}\smallskip
\endminipage\hfill
\minipage{0.245\linewidth}
  \includegraphics[width=\linewidth]{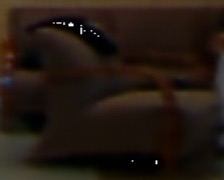}\smallskip
\endminipage\hfill
\minipage{0.245\linewidth}
  \includegraphics[width=\linewidth]{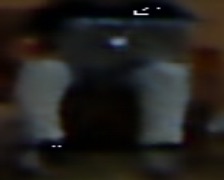}\smallskip
\endminipage\hfill

\caption {Regularizer significance examples. The first row depicts (from left to right)  (a) Hallucination without regularization, (b) Hallucination with regularization, and (c) Ground truth. The second row are zoomed version of the highlighted areas in (a)  and the third row are zoomed versions of the annotations in  (b) output.}
\label{UWregularizer}
\end{figure}

\textbf{Regularizer Network Significance:} Although the hallucinator network produces convincing RGB renditions from the depth images, the results still seem to display some visible discrepancies between the original RGB and hallucinated data. The hallucinator network seems to be concerned with reproducing the overall structure of the image and doesn't give much importance to color information. Moreover, as the hallucinator network is trained with a weighted smoothness constraint to ensure local smoothness the hallucinator network ignores smaller objects in the RGB  image. The regularizing autoencoder helps to overcome these shortcomings as seen in Fig. \ref{UWregularizer}. The regularizer helps to maintain the color information, texture, finer components of the images missed by hallucinator,  de-blurs the image and also removes irregularities. As seen in the Fig \ref{UWregularizer} and the color is better reconstructed by the regularizer. We can see the regularizer can reproduce smaller details such as an electrical socket on the wall or the Apple symbol on the computer. This ability of the regularizer makes it a non-trivial part of our experiment. Although the proposed two-stage method does comparably well, it violates one of the fundamental constraints we place on our method: online training. 

\subsubsection{AggConv Results}
\begin{figure}[!ht]
\vspace{10pt}
\centering
\includegraphics[width=.32\linewidth]{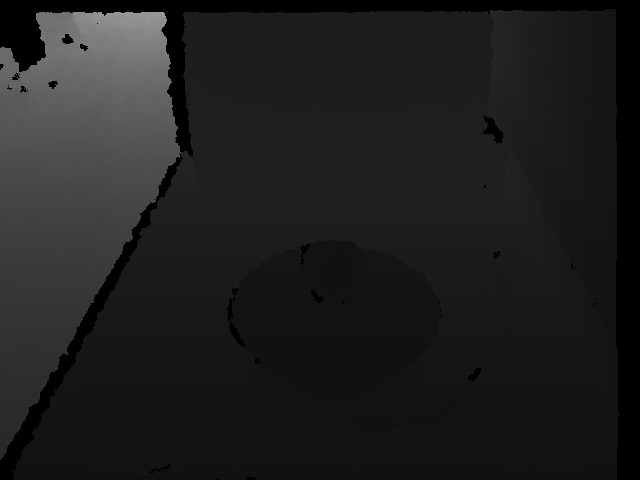}
\includegraphics[width=.32\linewidth]{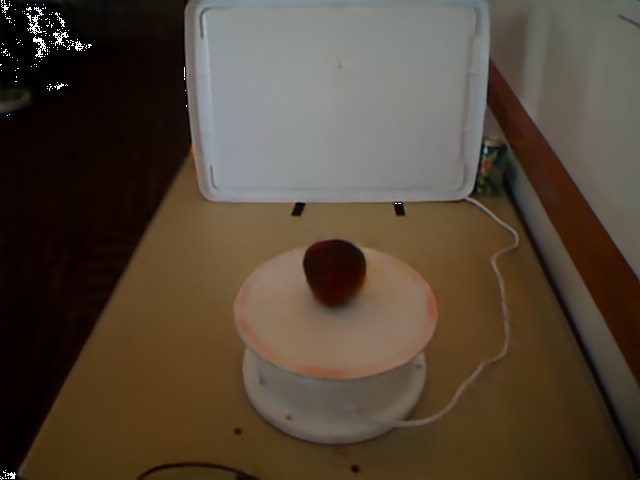}
\includegraphics[width=.32\linewidth]{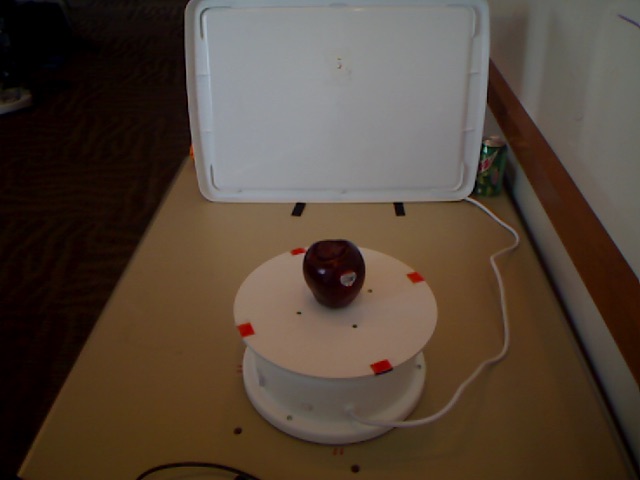}
\smallskip

\stackunder[3pt]{\includegraphics[width=0.32\linewidth]{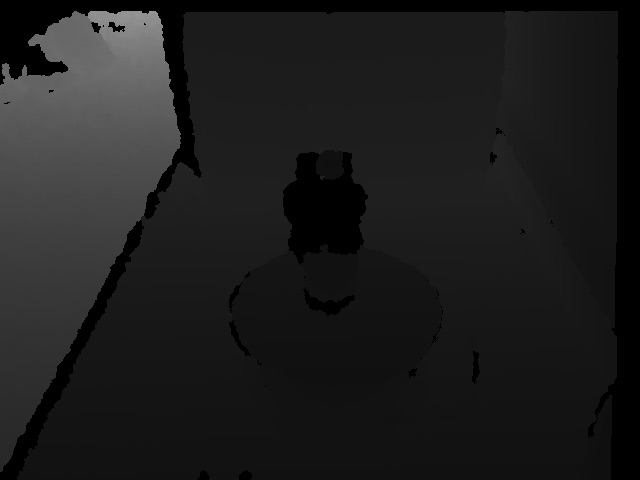}}
{(a) Depth}
\stackunder[3pt]{\includegraphics[width=.32\linewidth]{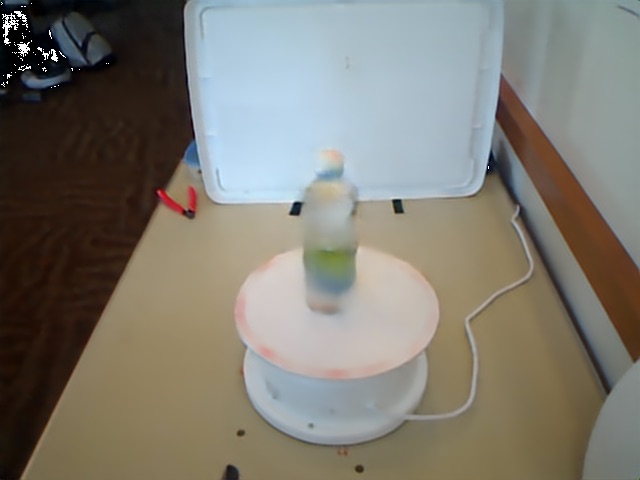}}{(b) Hallucinated}
\stackunder[3pt]{\includegraphics[width=.32\linewidth]{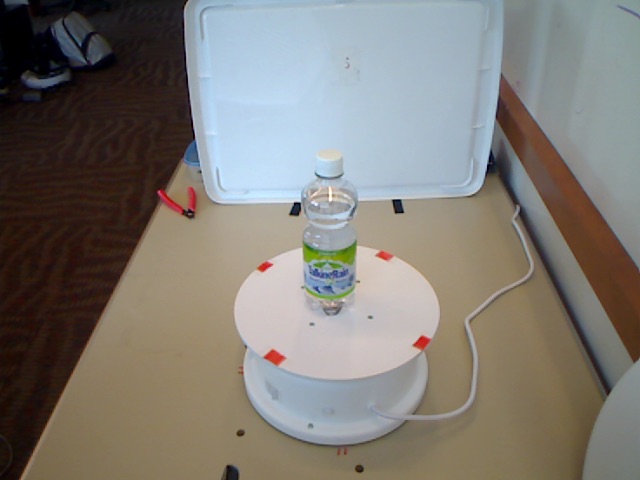}}{(c) RGB}
\caption{UWRGBD hallucination results using our proposed architecture with Aggregated convolutional blocks. (a) depth input image, (b) hallucination results, (c) is groundtruth. }
\label{agghaluw}
\end{figure}

\begin{figure}[!ht]
\centering
\includegraphics[width=.32\linewidth]{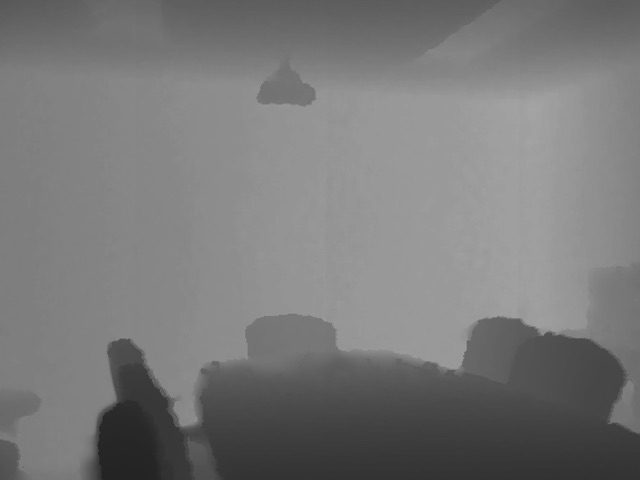}
\includegraphics[width=.32\linewidth]{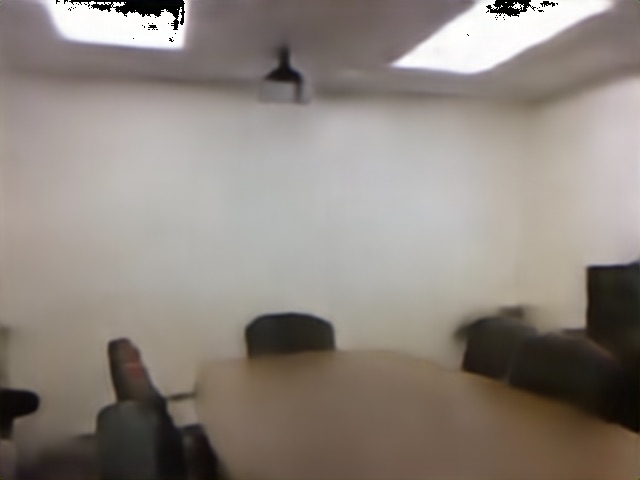}
\includegraphics[width=.32\linewidth]{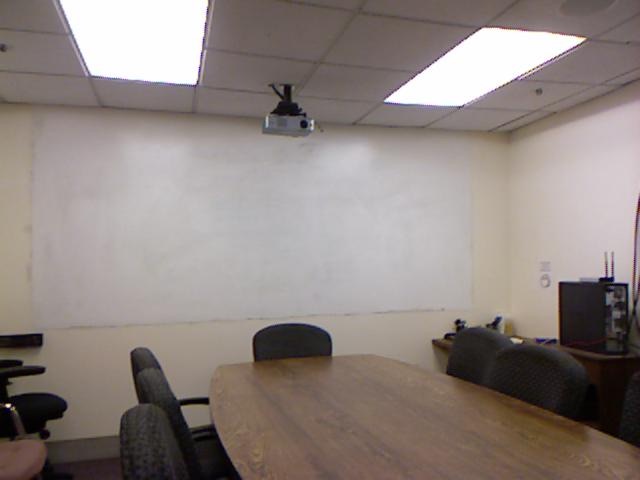}
\smallskip

\stackunder[3pt]{\includegraphics[width=0.32\linewidth]{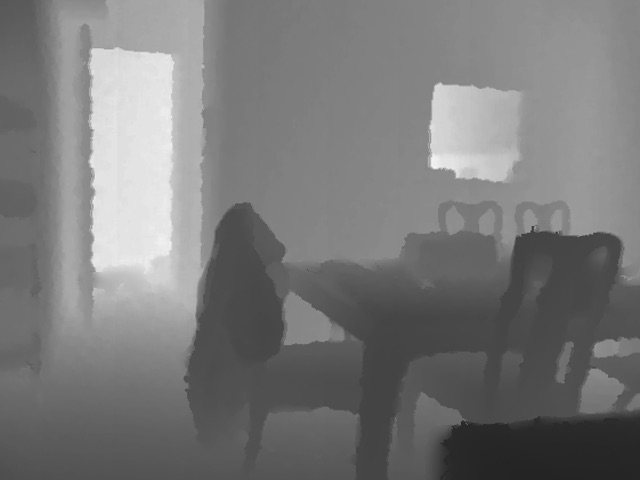}}
{(a) Depth}
\stackunder[3pt]{\includegraphics[width=.32\linewidth]{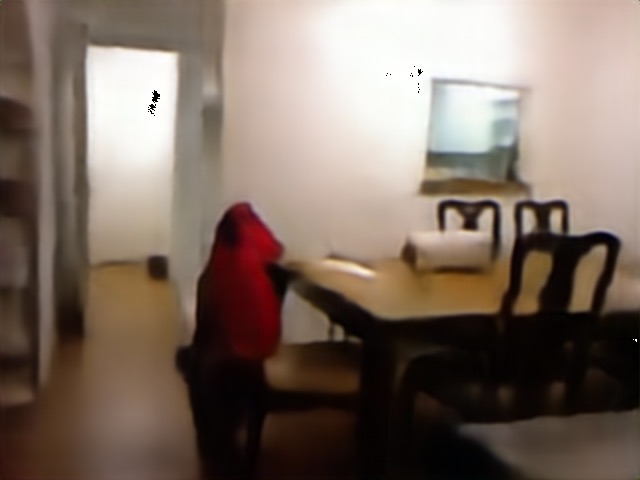}}{(b) Hallucinated}
\stackunder[3pt]{\includegraphics[width=.32\linewidth]{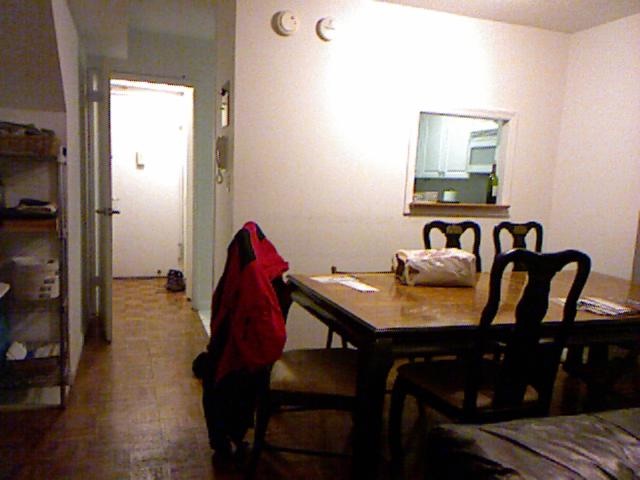}}{(c) RGB}
\caption{NYUD dataset hallucination results using AggConv blocks (our proposed architecture). (a) depth input image, (b) hallucination results, (c) is groundtruth. }
\label{agghalnyud}
\end{figure}

The advantage of using our architecture based on aggregated fields of view is evident from Fig. \ref{agghaluw} and \ref{agghalnyud}. The network works effectively in reproducing the color and structural information of the image in a single-stage pipeline without requiring the need for regularizer. It produces better results from a visual perspective compared to the GAN and linkNet architecture without violating any constraints unlike off-the-shelf models like linkNet. Thus, in an adverse scenario, an online trained network using our architecture would significantly mitigate the risk the system would be subjected to compared to a hallucination scheme using GAN or linkNet.

 Although visual inspection gives a general idea of the superior hallucination of our network, it is better judged with a quantitative metric that indicates how the test set hallucinations deviated from the ground truth. The table \ref{mapd_table} shows how mean absolute pixel difference of the test set and it provides more proof to the effectiveness of our architecture.

\begin{table}[!ht]
    \centering
    \begin{tabular}{|c|c|c|c|}
    \hline
    \multicolumn{4}{|c|}{\textbf{Mean Absolute Pixel Difference}}\\
    \hline
        Dataset & GAN  & LinkNet & AggConv(ours)  \\
        \hline
         NYUD &  137.57 & 10.76 & \textbf{5.96} \\
        UWRGBD & 134.155 & 3.39 & \textbf{2.36} \\
        \hline 
    \end{tabular}
    \caption{Mean absolute pixel difference indicates the average deviation of a pixel from it's true value. }
    \label{mapd_table}
\end{table}

{\bf Substituting the lost modality with hallucination helps:} The Table \ref{result_Single} provides evidence to the fact that the hallucinated images indeed captures some of the necessary RGB space information. We train a 51-way object classification network using \cite{alexnetpaper} architecture on the UWRGBD dataset and a 40-way Semantic segmentation task using \cite{seg} architecture on the NYUD dataset. We train the networks with RGB data and during the test phase, we replace it with the other modality images. Since we are trying to capture RGB space information a network trained on RGB data should be able to extract and use features from the hallucinated data and should perform better than the depth images. This can indeed be seen in Table \ref{result_Single}. (In Tables \ref{result_Double} and \ref{result_Single} the  RGB + Depth and RGB  columns respectively have been given for reference. It is the original system performance. Since we are considering lost RGB modality the performance comparison is between Depth and hallucinated modality).

\begin{table}[!ht]
    \centering
    \resizebox{\columnwidth}{!}
    {
    \begin{tabular}{|c|c|c|c|c|c|}
    \hline
    \multicolumn{6}{|c|}{Hallucinated Modality's Effectiveness}\\
    \hline
        \hline
    Task  & \vtop{\hbox{\strut Object}\hbox{\strut Classification}}  & \multicolumn{4}{c|}{Semantic Segmentation}   \\
    \hline
    Metric & \vtop{\hbox{\strut Total}\hbox{\strut Accuracy}}  & \vtop{\hbox{\strut Pixel}\hbox{\strut Accuracy}}  & \vtop{\hbox{\strut Mean}\hbox{\strut Accuracy}}  & \vtop{\hbox{\strut Mean}\hbox{\strut IoU}}   & \vtop{\hbox{\strut Freq.}\hbox{\strut IoU}}    \\ 
    \hline
    RGB & 96.67\% & 53.64\%  & 40.10\% & 30.13\% & 44.82\%  \\
    Depth & 2.12\% & 18.25\% & 12.95\% & 4.58\% & 10.23\%  \\
    GAN  & 27.48\% & 25.84\% & 18.20\% & 9.69\% & 18.04\% \\
    LinkNet & 29.19\% & 31.87\% & 19.87\% & 11.17\% & 22.07\% \\
    AggConv &  \textbf{51.14\%} & \textbf{35.78\%} & \textbf{22.22\%} & \textbf{13.42\%} & \textbf{25.33\%} \\
    \hline
    \end{tabular}
    }
    \caption{The table provides evidence for the effectiveness of using the hallucinated modality. AggConv is our method. }
    \label{result_Single}
\end{table}

{\bf Combining the working modality with the hallucinated modality maintains the overall system's performance:} The loss of the primary modality could be anticipated and as a countermeasure, the same task could be trained on other modalities, but that does not ensure good performance. For instance, a pipeline in a self-driving car could be trained for lane detection using the RGB camera data and as a back-up, a network for depth-based detection could be trained in the same way as well. The depth-based system would not perform as well, as that modality is not information-rich like the RGB modality for this task. We believe, in this case, the hallucinated data in combination with the depth data could be better than just having a depth data based back up. This can be seen well depicted in Table \ref{result_Double}. We use two-stream classification and segmentation networks to show the benefits of incorporating the hallucinated modality with the depth modality. The original system is trained with RGB and depth data. Both classification and segmentation tasks perform much better with hallucinated and depth modalities together than just having depth. There is an increase of approximately 50\% classification accuracy and 2.5\% mean IoU score for segmentation task which is a significant increase for semantic segmentation. This result validates our claim that data can be hallucinated and be used along with the lower dimensional data to reduce the risk. The performance is comparable to the performance of the original system. Thus in the case of a lost modality, hallucinated data can be helpful. 

\begin{table}[!ht]
    \centering
    \resizebox{\columnwidth}{!}
    {
    \begin{tabular}{|c|c|c|c|c|c|}
    \hline
    \multicolumn{6}{|c|}{Hallucinated Modality Reduces Risk}\\
    \hline
        \hline
    Task  & \vtop{\hbox{\strut Object}\hbox{\strut Classification}}  & \multicolumn{4}{c|}{Semantic Segmentation}   \\
    \hline
    Metric & \vtop{\hbox{\strut Total}\hbox{\strut Accuracy}}  & \vtop{\hbox{\strut Pixel}\hbox{\strut Accuracy}}  & \vtop{\hbox{\strut Mean}\hbox{\strut Accuracy}}  & \vtop{\hbox{\strut Mean}\hbox{\strut IoU}}   & \vtop{\hbox{\strut Freq.}\hbox{\strut IoU}}    \\ 
    \hline
    % \multicolumn{6}{||c||}{\textbf{Setting A}} \\   
    % \hline
    RGB+Depth & 97.78\% & 55.52\% & 42.30\%  & 32.08\% & 46.60\%   \\
    Depth & 53.15\% & 50.53\% & 35.73\% & 26.05\% & 40.67\%  \\
    GAN+Depth & 86.15\% & 52.45\% & 37.19\% & 27.46\% & 42.57\% \\
    LinkNet+Depth & 88.01\% & 52.03\% & 38.15\% & 28.02\% & 42.33\% \\
    AggConv+Depth &  \textbf{92.37\%} & \textbf{52.95\%} & \textbf{38.51\%} & \textbf{28.61\%} & \textbf{43.32\%} \\
    \hline
    \end{tabular}
    }
    \caption{This table shows the benefits of incorporating the data from the hallucinated modality with the depth modality when the RGB modality is lost. }%It can be seen that the risk to the system is reduced.}
    \label{result_Double}
\end{table}

{\bf Incorporating the hallucinated modality while all others are working enhances the overall system's performance:} An added advantage that we observed from the hallucinated data is that it can be incorporated into the original system to improve the system performance. The hallucinated data captures the space between the depth modality and the RGB modality and combining it with the original system results in gains in performance. Table \ref{result_Triple} provides evidence for the same thus proving hallucinated data aides in enhancing the performance existing system. Both classification and segmentation task benefit from the added modality with segmentation gaining as much as 2.5\% on mean IoU score.

\begin{table}[!ht]
    \centering
    \resizebox{\columnwidth}{!}
    {
    \begin{tabular}{|c|c|c|c|c|c|}
    \hline
    \multicolumn{6}{|c|}{Hallucinated Modality Enhances.}\\
    \hline
        \hline
    Task  & \vtop{\hbox{\strut Object}\hbox{\strut Classification}}  & \multicolumn{4}{c|}{Semantic Segmentation}   \\
    \hline
    Metric & \vtop{\hbox{\strut Total}\hbox{\strut Accuracy}}  & \vtop{\hbox{\strut Pixel}\hbox{\strut Accuracy}}  & \vtop{\hbox{\strut Mean}\hbox{\strut Accuracy}}  & \vtop{\hbox{\strut Mean}\hbox{\strut IoU}}   & \vtop{\hbox{\strut Freq.}\hbox{\strut IoU}}    \\ 
    \hline
    % \multicolumn{6}{||c||}{\textbf{Setting A}} \\   
    % \hline
    RGB+Depth & 97.78\% & 55.52\% & 42.30\%  & 32.08\% & 46.60\%   \\
    RGB+Depth+GAN & \textbf{98.83\%} & 57.45\% & 44.41\% & 34.35\% & 48.83\%  \\
    RGB+Depth+LinkNet & 97.44\% & 57.12\% & 44.06\% & 33.92\% & 48.54\% \\
    RGB+Depth+AggConv &  98.12\% & \textbf{57.53\%} & \textbf{44.75\%} & \textbf{34.41\%} & \textbf{48.85\%} \\
    \hline
    \end{tabular}
    }
    \caption{The  hallucinated modality can be incorporated with the fully functioning system to enhance performance. RGB +  Depth + AggConv is our method. }
    \label{result_Triple}
\end{table}

\subsection{Discussion}
The proposed method is intended for systems that are assumed to be working well and for a reasonable time before the adverse event happening. This is an essential assumption as the training data for the hallucination scheme is generated during the normal working condition on the robot or autonomous system. 

An object's information such as color that cannot be obtained from the lower-dimensional modality, in the depth modality is obtained using the correlation between that information and the neighborhood of the object of interest. This correlation is learned from the explicit relationship that exists in the training data. When this relationship no longer holds, during inference the model still predicts from the relationship that is in memory. 

To better explain this consider the hallucination of a wall from a depth image. Now, the best answer for assigning the color to this wall would be to assign it the average of all the colors seen in the training set. With our architecture, the model can use the neighbors of the wall to predict it's color and texture. If the neighborhood is associated with lamps, books and writing desks, the model could make an interpretation that it' is a study room and assign the color based on training observations it has seen. Suppose, the wall color is changed all of a sudden a robot running the trained hallucination model will fail in predicting the new color and texture. Thus successful hallucination is limited by the constancy of these relationships between the pixel and its neighbors. However,  given enough time and training data, the model should be able to learn new relationships.

\begin{figure}[!ht]
\centering
\includegraphics[width=.32\linewidth]{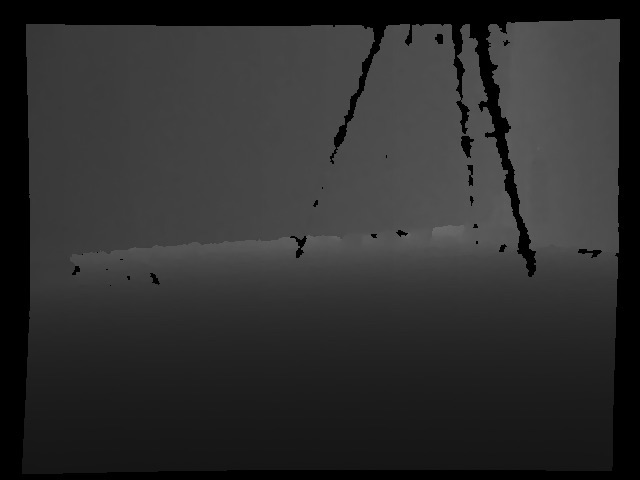}
\includegraphics[width=.32\linewidth]{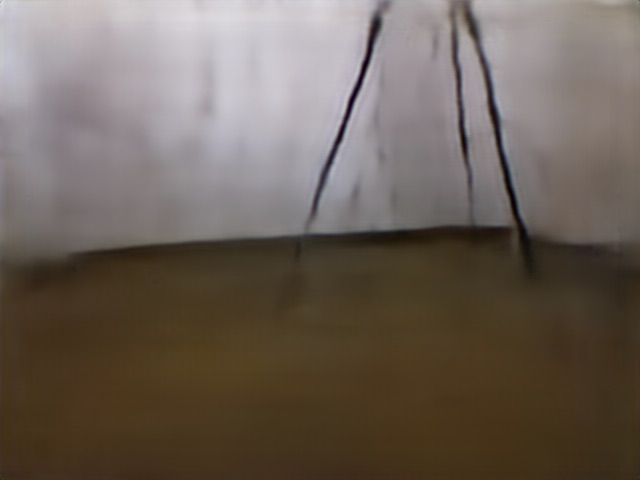}
\includegraphics[width=.32\linewidth]{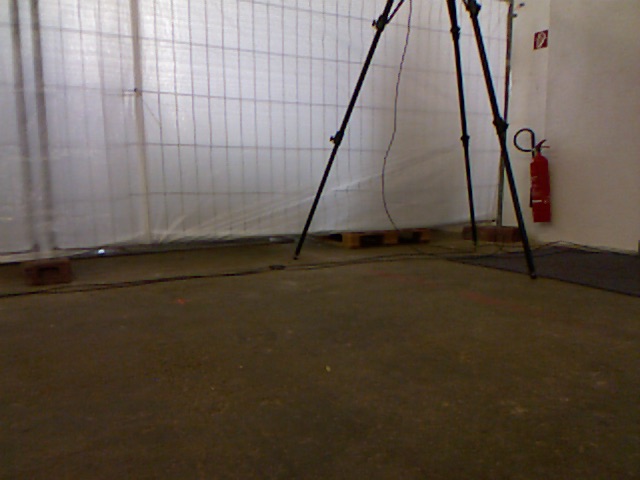}

\smallskip

\stackunder[3pt]{\includegraphics[width=0.32\linewidth]{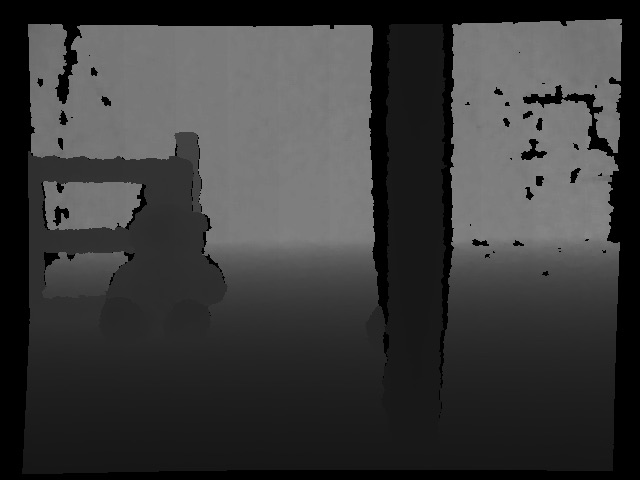}}
{(a) Depth}
\stackunder[3pt]{\includegraphics[width=.32\linewidth]{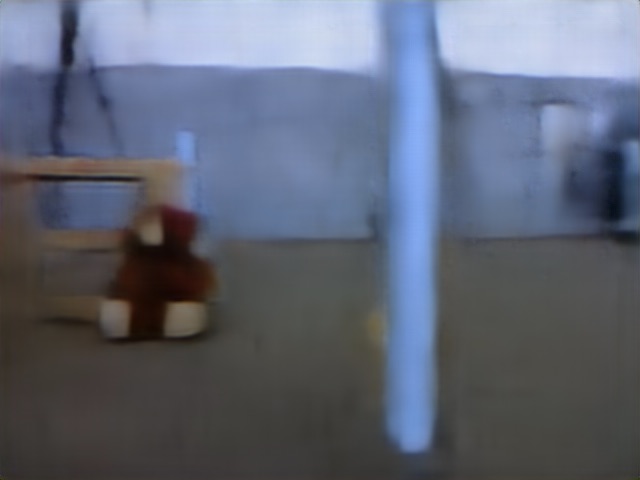}}{(b) Hallucinated}
\stackunder[3pt]{\includegraphics[width=.32\linewidth]{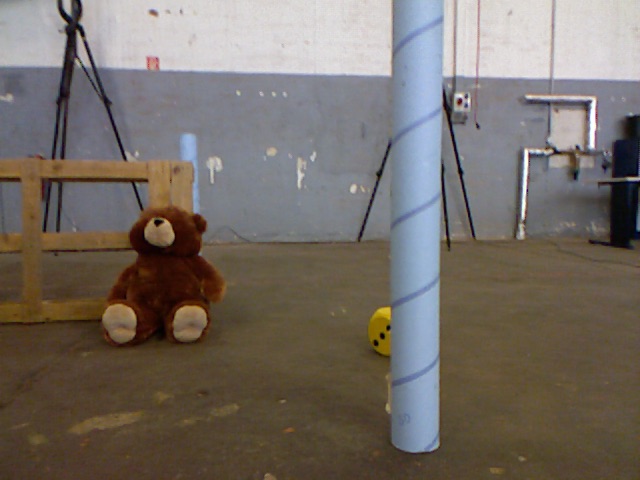}}{(c) RGB}
\caption{The NYUD trained model fine-tuned with TUM. (a) depth input, (b) hallucination output, (c) groundtruth. }
\label{trained}
\end{figure}

To further demonstrate the trained hallucination model is able to generalize well, we finetuned the NYUD trained one with data from the TUM dataset\cite{tum}. In particular, we use the dataset under "robot slam" category to do this. The sequences "fr2/pioneer360", "fr2/pioneer\_slam" and "fr2/pioneer\_slam3" were used as training dataset while "fr2/pioneer\_slam2" is used as the testing dataset. The results shown in Fig. \ref{trained}  are from test set.  There are in all 6000 and odd images in the training set and 2000 and odd images in the test set, hence we decided to fine-tune the NYUD dataset trained model instead of training it from scratch. It was trained for 8 epochs with a batch size of 7. The hallucination model does a pretty good job in preserving the overall information of the scene even with a small data sequence, thus indicating the ability of our model to quickly learn and adapt to new and changing environments (see video attached). 
 
\section{Conclusion}

We bring to light the importance of hallucination in multimodal systems and the challenges in hallucinating from low to high dimension modality. We describe a common adverse scenario in autonomous systems, which is the loss of a data modality and present a method to hallucinate data from the existing modality by capturing a non-linear mapping between the data spaces. We introduce a field of view aggregating convolutional block (AggConv and AggTrConv) compare our proposed hallucination architecture with state of the art networks re-purposed for this task of hallucination. We provide qualitative and quantitative results as evidence and further validate our claim on two vision tasks (classification, segmentation) and show that the hallucinated modality does reduce the risk to the system that arises due to modality loss. We have made our implementation and data samples for experimentation publicly available for future research.

\noindent{\bf Acknowledgment:} 
%This work is partially supported byNSF CAREER IIS-1750082 and NRI-1925403. We acknowledge NVIDIA for the donation of GPUs.
The National Science Foundation under Robust Intelligence program (\#1750082), and National Robotics Initiative program (\#1925403), AWS MLRA, and GPU donations from NVIDIA are gratefully acknowledged.
% if have a single appendix:
%\appendix[Proof of the Zonklar Equations]
% or
%\appendix  % for no appendix heading
% do not use \section anymore after \appendix, only \section*
% is possibly needed

% use appendices with more than one appendix
% then use \section to start each appendix
% you must declare a \section before using any
% \subsection or using \label (\appendices by itself
% starts a section numbered zero.)

% Can use something like this to put references on a page
% by themselves when using endfloat and the captionsoff option.
\ifCLASSOPTIONcaptionsoff
  \newpage
\fi

\bibliographystyle{ieeetr}
\bibliography{ref}
% that's all folks
\end{document}